# Implementing a Real-Time, YOLOv5 based Social Distancing Measuring System for Covid-19

Narayana Darapaneni[1], Shrawan Kumar[2], Selvarangan Krishnan[3], Hemalatha K[4], Arunkumar Rajagopal[5], Nagendra[6], and Anwesh Reddy Paduri[7]

[1] Northwestern University/Great Learning, Evanston, US
[2-7] Great Learning, Bangalore, India

anwesh@greatlearning.in

**Abstract.** The purpose of this work is, to provide a YOLOv5 deep learning-based social distance monitoring framework using an overhead view perspective. In addition, we have developed a custom defined model "YOLOv5 modified CSP (Cross Stage Partial Network)" and assessed the performance on COCO and Visdrone dataset with and without transfer learning. Our findings show that the developed model successfully identifies the individual who violates the social distances. The accuracy of 81.7% for the modified bottleneck CSP without transfer learning is observed on COCO dataset after training the model for 300 epochs whereas for the same epochs, the default YOLOv5 model is attaining 80.1% accuracy with transfer learning. This shows an improvement in accuracy by our modified bottleneck CSP model. For the Visdrone dataset, we are able to achieve an accuracy of upto 56.5% for certain classes and especially an accuracy of ~40% for people and pedestrians with transfer learning using the default YOLOv5s model for 30 epochs. While the modified bottleneck CSP is able to perform slightly better than the default model with an accuracy score of upto 58.1% for certain classes and an accuracy of ~40.4% for people and pedestrians.

## 1 INTRODUCTION

The coronavirus disease was reported in December 2019 in China. Soon the virus caused a global outbreak, and the World Health Organization (WHO) announced the situation as pandemic [1]. The data published by WHO on 4th November 2021 confirms 247.96 million infected people and a scary number of 5,020,204 deaths globally.

On 8 July 2020, the WHO announced "There is emerging evidence that COVID-19 is an airborne disease that can be spread by tiny particles suspended in the air after people talk or breathe, especially in crowded, closed environments or poorly ventilated settings"

According to WHO, the minimum distance between individuals must be at least 6 feet to ensure adequate social distance among the people.

The main challenges are attaining a high level of accuracy, lighting conditions, occlusion, and real-time performance. The key goals of this work are as follows:

- To present YOLOv5 deep learning-based social distance monitoring tool using an overhead view perspective.
- To deploy pre-trained YOLOv5 for person detection and computing their bounding box centroids. In addition, a transfer learning method is applied to improve the performance of the model trained on COCO and Visdrone dataset.
- To assess the performance of a custom defined model "YOLOv5 modified CSP (Cross Stage Partial Network)" on COCO and Visdrone dataset without transfer learning.
- In order to track the social distance between individuals, the Euclidean distance is used to approximate the distance between each pair of the centroid of the bounding box detected. In addition, a social distance violation threshold is specified using a pixel to distance estimation.
- To assess the social distancing model performance on an overhead data set.

The rest of the work discussed in this paper is structured as follows. The related work is presented in Section 2. The salient features of the dataset which we used to train and validate is presented in Section 3. A deep learning-based social distance monitoring framework has been presented in Section 4. The detailed analysis of output results and performance evaluation of the model with and without transfer learning is also illustrated in this Section 5. The future scope and challenges are given in the Section 6. The conclusion of the given work with potential future plans is provided in Section 7.

## 2 RELATED WORKS

Convolutional Neural Networks have played a very crucial role in complex object classification and feature extraction, including human detection. Over the past few decades, convolutional neural networks (CNN), faster region-based CNN (Faster RCNN), and region-based CNN (RCNN) used region proposal techniques to produce the objectness score prior to its classification and later created the bounding boxes around the interested objects for visualization and statistical analysis. With the development of GPUs, faster CPUs, and extended memory capacities, the researchers were able to build the CNN systems with high accuracy and fast detection compared to conventional models. Even though the above-mentioned methods are efficient but endure in terms of detection speed, long time training and achieving better accuracy, there are still remaining issues to be solved.

Since all these CNNs based model approaches classification, YOLO (You Only Look Once) considered a different approach and used a regression-based method to dimensionally split the bounding boxes and translate their class probabilities. In YOLO, the framework efficiently splits the image into multiple portions representing bounding boxes with the probability scores for class for each portion to consider as an object. YOLO provides significant improvements in terms of speed at the cost of reduced efficiency as well as an object detector module that exhibits powerful generalization capabilities to represent the whole image.

After an object is detected, classification methods can be used to identify a human on the basis of motion-based features, shape, or texture. In shape-based methods, shape related information about moving regions are defined to detect the human. Due to limitations in standard template matching schemes, this method performs not so great which is further augmented by incorporating part-based template matching approach. Dalal et al. [51] suggested texture-based schemes such as histograms of oriented gradient, which makes use of higher dimensional features along with the support vector machine to detect the humans.

Human identification in image or video sequences is a very crucial part in the field of computer vision and object detection and it is an important subbranch in this field. Although many researchers have worked on human action recognition and human detection, it is mostly either limited to indoor applications or faces accuracy issues under outdoor challenging conditions which is not limited to lighting conditions. Other research works employ manual-tuning methodologies to classify/identify people's activities, however, limited functionality has been an issue.

Recent research shows that gait and face recognition techniques can be used for further identification of humans in surveillance video. However, tracking and detection of people specially under a crowd is difficult at times due to full or partial occlusion problems. Leibe et al. [17] came up with a solution based on trajectory estimation while Andriluka et al. [50] proposed tracklet-based detectors as a solution in detecting partially occluded people.

With many social applications, crowd counting emerged as a key area of research. Eshel et al., [13] focused on person count and crowd detection by suggesting multiple height homographies for head top detection and overcame the occlusions problem related to video surveillance applications. Chen et al. [52] developed an application for electronic advertising using the concept of crowd counting. For a similar application, Chih-Wen et al. [53] proposed a vision-based people counting model. Following the footsteps, Yao et al. [54] captured inputs from stationary cameras in order to carry out background subtraction in order to train the model for the foreground shape and appearance of the crowd in videos.

Rahim A et al. [55] proposed a framework which utilizes the YOLOv4 model for real-time object detection. They also proposed the social distance measuring approach in their YOLOv4 model framework to specify the risk factor based on the calculated distance and safety distance violations. In this model they introduced a single motionless time of flight (ToF) camera to capture the video sequence with various lighting conditions.

Mahdi Rezaei et al. [56] proposed the framework which uses YOLOv4 based Deep Neural Network (DNN) model to automate human detection in crowded places both indoor and outdoor environments using common CCTV cameras. They used the DNN model in combination with an adapted inverse perspective mapping (IPM) technique and SORT tracking algorithm to detect people and social distance monitoring.

Imran Ahmed et al. [57] proposed a deep learning platform based on YOLOv3 model in detecting humans and correspondingly track their social distance using an overhead perspective. The detection algorithm uses a pre-trained model that is connected to an extra trained layer using an overhead data set. The detection model detects humans using bounding box information. The bounding box centroid values using Euclidean distance used to evaluate the pairwise distances between humans to evaluate social distance violations.

Sergio Saponara et al. [58] proposed a deep learning model using YOLOv2 to detect and track people in indoor and outdoor scenarios. The proposed approach required the images acquired through thermal cameras to establish a complete AI system for people tracking, social distance classification, and body temperature monitoring.

A few other research works propose new loss functions in order to effectively address the problem of crowded detections. For example, Occlusion-aware R-CNN suggests aggregation loss in order to enforce proposals to be as close to the corresponding objects and reduces the internal region distances of proposals related to the same objects. Repulsion Loss adds an extra penalty to proposals intertwined with several ground truths. Furthermore, advanced NMS strategies are put forward to reduce the crowdedness issues to some degree, but they still use IoU as the metric to calculate the difference between detected objects, which results in limiting the performance on recognizing highly overlapped instances in crowded boxes.

# 3 SALIENT FEATURES OF DATASET USED FOR TRAINING AND VALIDATION

## 3.1 Visdrone:

- AISKYEYE team collected the VisDrone2019 dataset at the Lab of Machine Learning and Data Mining, Tianjin University, China.
- The Visdrone dataset contains 288 video clips made up of 261,908 frames and 10,209 images.
- The images and videos have been captured by various drone-mounted cameras, covering a wide range of aspects including location (taken from 14 different cities separated by thousands of kilometers in China), environment (urban and country), objects (pedestrian, vehicles, bicycles, etc.), and density (sparse and crowded scenes).
- In object detection tasks, we focus on ten object categories of interest including pedestrian, person, car, van, bicycle, awning-tricycle, bus, truck, motor, and tricycle.

## 3.2 COCO Dataset:

- COCO, short for Common Objects in Context, is a large image recognition / classification, object detection, segmentation, and captioning dataset.

- The COCO Dataset has 121,408 images
- The COCO Dataset has 883,331 object annotations
- The COCO Dataset has 80 classes
- The COCO Dataset median image ratio is 640 x 480

### 3.3 Data collected for Validation:

Our team has collected and curated over 100s of videos in order to test the social distancing output indicators on the data using YOLOv5. The collected videos have the following features:

- Day or night capture
- Shade or sunlight capture
- Crowd – High or low
- Different types of lights during night capture
- Various angles of capture
- Static or dynamic capture – Some videos were captured from a static position while some were captured from dynamic (moving) position.
- Normal or speed up capture – Some videos are in normal motion while some are in 2X motion.

## 4 METHODOLOGY

This study aims to implement the first working model of YOLOv5 based social distancing model using YOLOv5s, YOLOv5s6, and YOLOv5s6 modified bottleneck CSP trained on COCO dataset.

In addition, we have made modifications to existing YOLOv5 architecture and came up with a new architecture named YOLOv5s6 modified backbone CSP.

We have also trained, validated, and detected the model performances on two datasets namely COCO and Visdrone.

### 4.1 Architecture of YOLOv5:

Anchor boxes consist of a set of predefined bounding boxes having certain height and width. These boxes are created to capture the scale and aspect ratio of the object classes to be detected and are chosen based on training dataset object sizes.
**Backbone: Feature Extraction**- It's a deep neural network (DNN) composed of convolution layers. Backbone is used to extract the essential features and therefore selection of the backbone is critical to ensure performance of object detection. Usually, to train the backbone, pre-trained neural networks are used.

- **Variants** - VGG, ResNet, SpineNet, CSPResNex50, CSPDarkNet53, EfficientNet, Darknet53, Inception/GoogleNet.

- **Activation** - Non-Linear Transformation. ReLu, LeakyReLu, ReLu6, SELU, Swish, Mish, Param-ReLu.

**Neck: Collect Feature Maps**- Neck is used mainly for collecting feature maps present in different stages. Neck consists of multiple top-down and bottom-up paths. Spatial Pyramid Pooling (SPP) Layer enables us to generate fixed size features irrespective of the size of our feature maps. Pooling layers such as Max Pooling are used to generate fixed size features. PaNet has introduced an architecture that allows improved propagation of layer information from bottom to top or top to bottom. The neck components flow up and down among layers and mainly connect a few layers at the end of the convolutional network. A series of layers to mix and combine image features to pass them forward to predictions.

- **Variants** - SPP, ASPP, PAN, FPN, BiFPN, ASFF, RFB, SFAM, NAS-FPN
- **Regularization** - DropOut, L1, L2, DropPath, SpatialDo, DropBlock

**Head: Predicting Class Probability**- For one stage detector, head is used to perform dense prediction. The final prediction is same as the dense prediction which consist of a vector having the coordinates (center, height, width) of the predicted bounding box and the confidence score of the prediction and the label.

- **Dense** - YOLO, SSD, RetinaNet, RPN, CornerNet, MatrixNet, R-FCN, FCOS

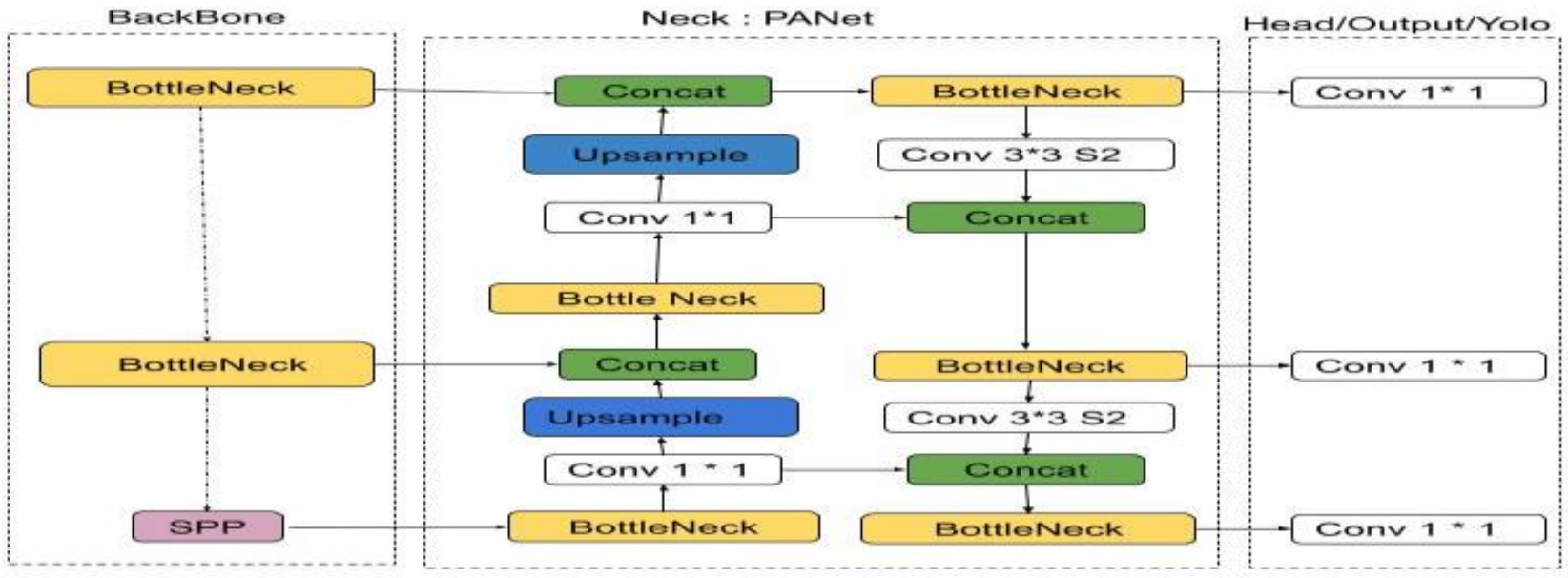

**Fig. 1.** YOLOv5 Architecture

Where,

SPP is Spatial Pyramid Pooling
Conv is convolution Layer
Concat is Concatenate function
PANet is a Path Aggregation Network.

## 4.2 Understanding the working of YOLOv5:

For multi-object detection, YOLO works well specially when one grid cell is associated with each object. Anchor box is helpful in case of overlap between objects which enables in detecting multiple objects in one grid cell.

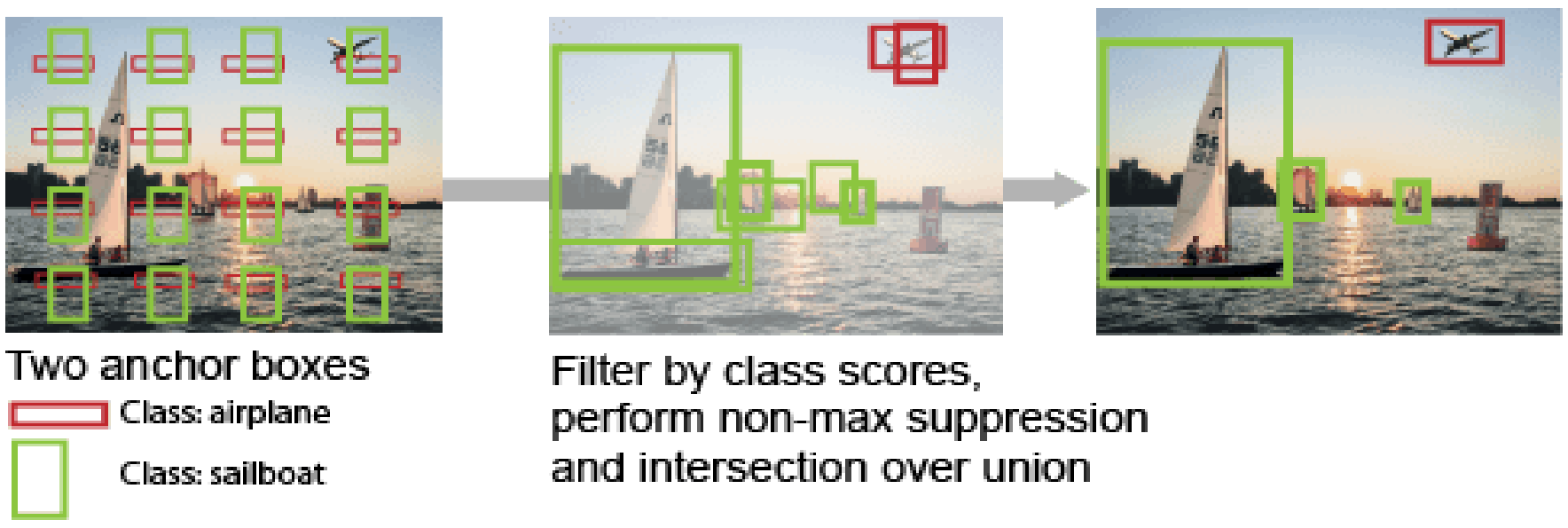

**Fig. 2.** Anchor box representation

### 4.3 YOLOv5 model variations

**a. YOLOv5s6 vs YOLOv5s6_modified:.** The dataset used is COCO128 and trained for 300 epochs. It is trained in 80 classes. The model depth multiple (depth_multiple) is 0.33. The layer channel multiple (width_multiple) is 0.50. The model uses four output layers P3, P4, P5, P6 respectively with strides 8, 16, 32, 64. The anchors vary from YOLOv5s file. YOLOv5s6 has four anchors. It has four anchors. We can supply anchors in our models, but the auto Anchor runs before training to ensure our anchors are a good fit for our data. If they are not, new anchors are evolved and attached to the model automatically (model.pt file). No change is required yaml file. If anchors are modified, it will get displayed to us and ask us to update the yaml file for future use. It has 3 main layers such as Backbone, Neck, and output. The neck and output layers have convolution, upsample, concat and C3. The backbone has conv, SPPF and the major difference is C3 (explain c3?) used in YOLOv5s6 whereas (explain bottleneck CSP?) bottleneck CSP in YOLOv5s6_modified version.

**b. YOLOv5s vs YOLOv5s6_modified:.**

. The dataset used is Visdrown and trained for 30 epochs. It is trained in 10 classes such as Pedestrian, People, Bicycle, Car, Van, Truck, Tricycle, Awning-tricycle, Bus and Motor. The model depth multiple (depth_multiple) is 0.33. The layer channel multiple (width_multiple) is 0.50. The YOLOv5s model uses three output layers P3, P4, P5 respectively with strides 8, 16, 32. It has three anchors. The YOLOv5s6_modified model uses four output layers P3, P4, P5, P6 respectively with strides 8, 16, 32, 64. It has four anchors. It has 3 main layers such as Backbone, Neck, and output. The neck and output layers have convolution, upsample, concat and C3. The backbone has conv, SPPF and the only difference is C3 in YOLOv5s and bottleneck CSP in YOLOv5s6_modified version.

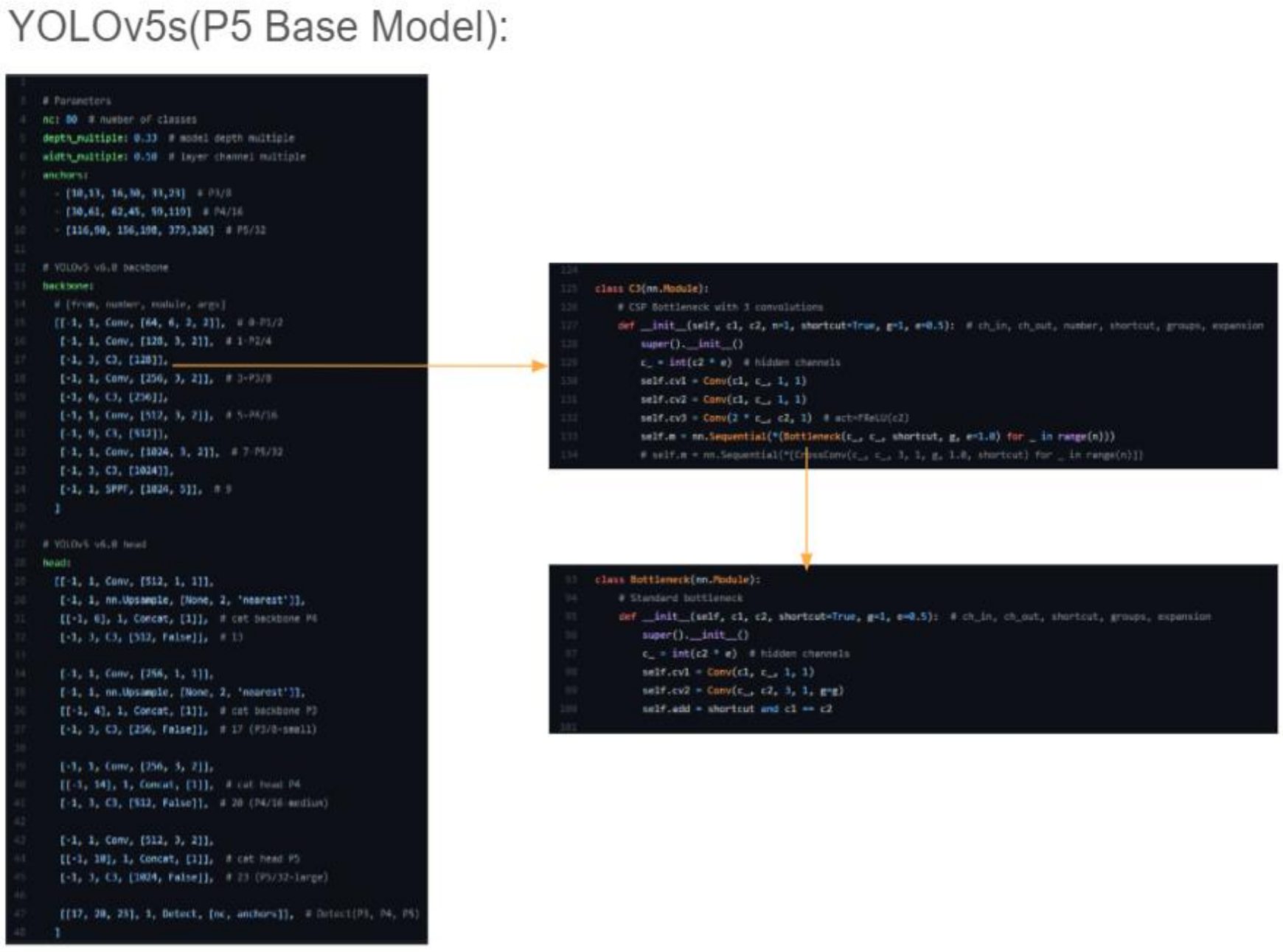

**Fig. 3.** YOLOv5s Base Model

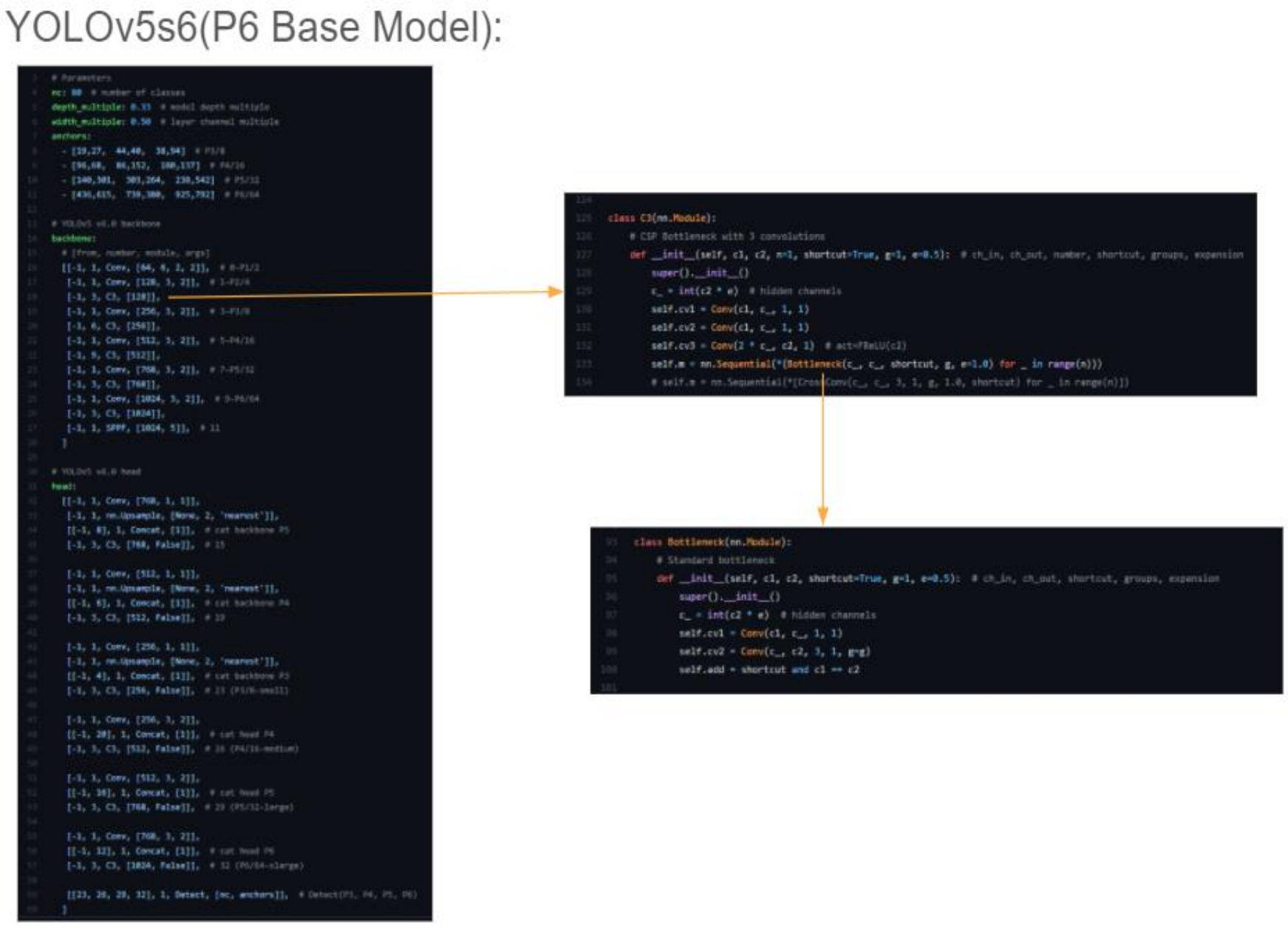

**Fig. 4.** YOLOv5s6 Base Model

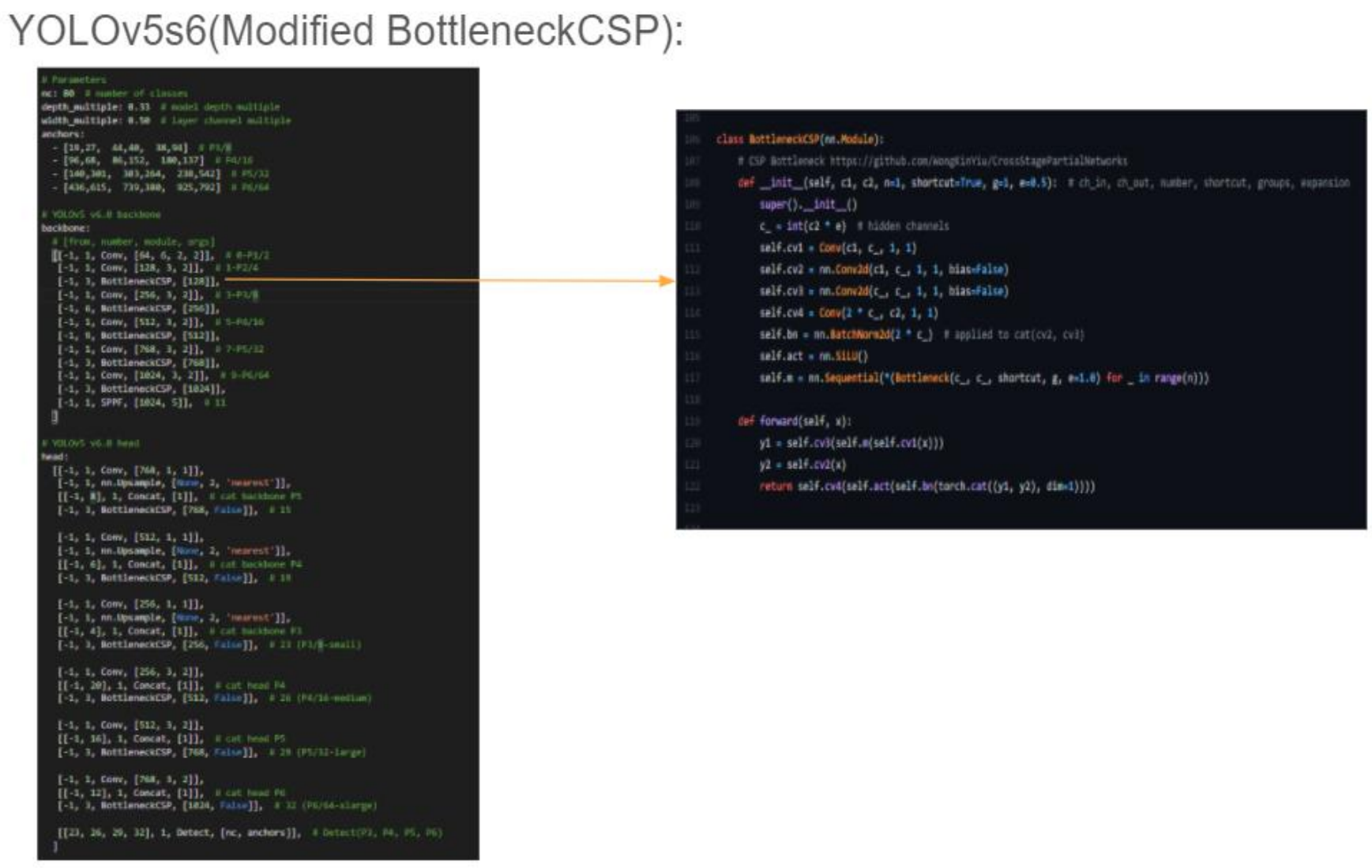

**Fig. 5.** YOLOv5s6_Modified Bottleneck CSP

## 4.4 Social Distance Monitoring:

To the best of our knowledge, we couldn't find any YOLOv5 social distancing algorithm successfully implemented and also using YOLOv5s, YOLOv5s6, and YOLOv5s6 modified bottleneck CSP on COCO dataset. This is a major gap which we wanted to address in our paper. In order to execute, we took the YOLOv5 based model and incorporated a detailed social distancing algorithm to successfully capture "Low Risk", "Medium Risk", "High Risk".

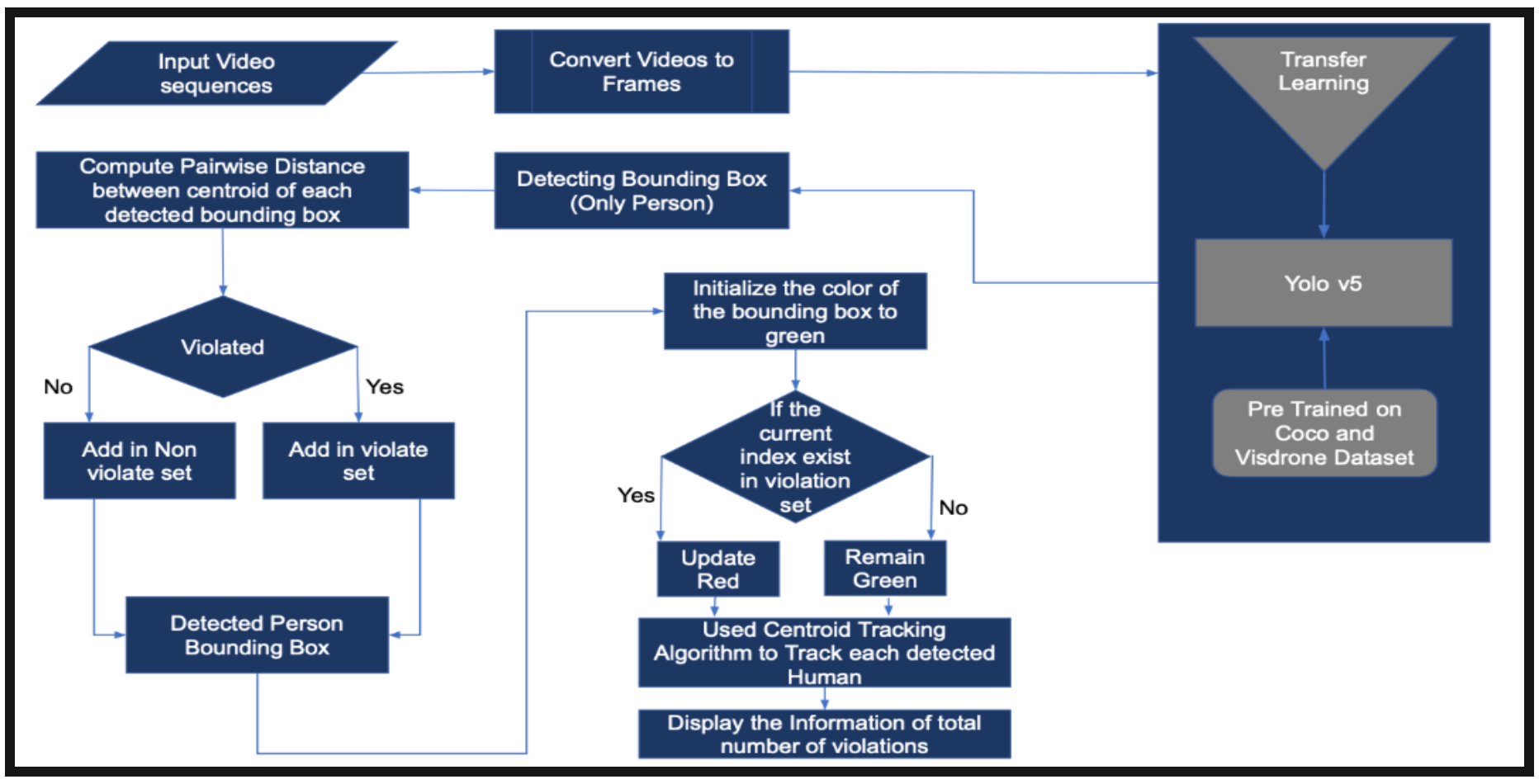

**Fig. 6.** The above flow chart describes the steps involved for people detection and social distancing classification.

Input a streaming video from a camera which contains people. Convert the videos into frames.

- Applying YOLOv5 object detector to detect people in each frame. The YOLO Architecture can be seen below.
- Check the number of persons that are in the Frames.
- Compute the distance between the centroid of the bounding boxes which are enclosed to the detected people.
- If violated distance above 50 cm (Configurable) add them in the violated set else in non-violate set.
- Initialize the color of the bounding box into green & loop through the index to check if it is available in the violation set. Update the violated index & non violated remain in green color.
- Finally, the algorithm will decide for safe or unsafe social distancing based on the number of persons and the measured distance between the centroid of bounding boxes

# 5 MODEL TRAINING & EXPERIMENTAL RESULTS

There is an existing implementation using YOLOV5 in github however the model is incomplete and also doesn't support additional architectures due to which is not published. We used this as reference and built a YOLOv5 social distancing model which is capable of working on multiple architectures used in the YOLOv5 model as well on our custom architecture (modified backbone CSP). By doing this, we have successfully implemented the first working social distancing model based on YOLOv5 and other supporting architectures based on YOLOv5.

The below benchmark comparisons are captured in the previous sections and the same can be considered for this section as well since the section heading/coverage is duplicate in nature.

## 5.1 Part1: Implementation of Social Distancing on YOLOv5 Models

Our key objective is to implement a working YOLOv5 based social distancing model using YOLOv5s, YOLOv5s6, and YOLOv5s6 modified bottleneck CSP architecture. There is no existing social distancing working model based on the above YOLOv5 architectures. From the screenshots below the model's sample output can be seen which clearly indicates the risk category based on centroid distance calculations.

**Parameters for High, Medium, and Low risk:**

**High Risk:** Distance between people less than 200 units.
**Medium Risk:** Distance between people between 200 - 250 units.

**Low Risk:** Distance between people more than 250 units.

Sample Outputs:

Video 1:

**Fig. 7.** Input Video is given to the YOLOv5 Social Distancing model

**Fig. 8.** Output Video from the YOLOv5 Social Distancing model

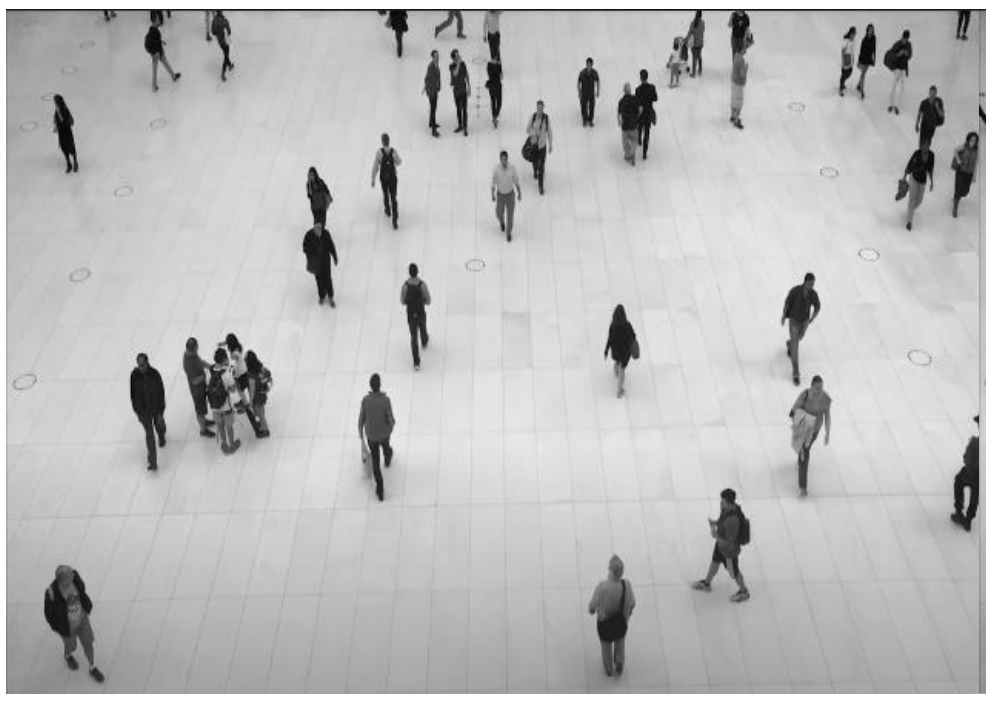

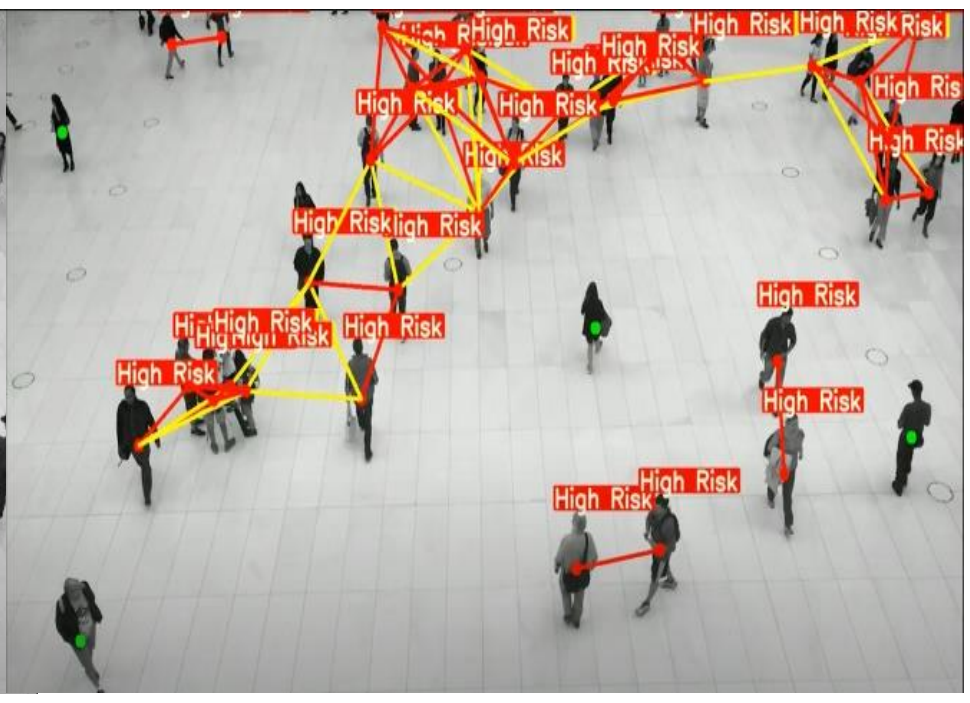

Video 2:

**Fig. 9.** Input Video is given to the YOLOv5 Social Distancing model

**Fig. 10.** Output Video from the YOLOv5 Social Distancing model

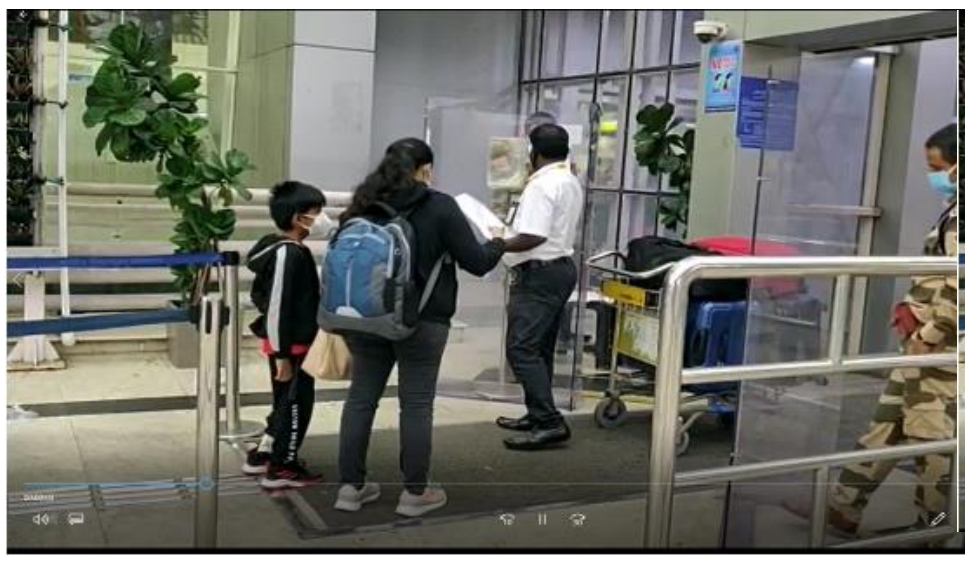

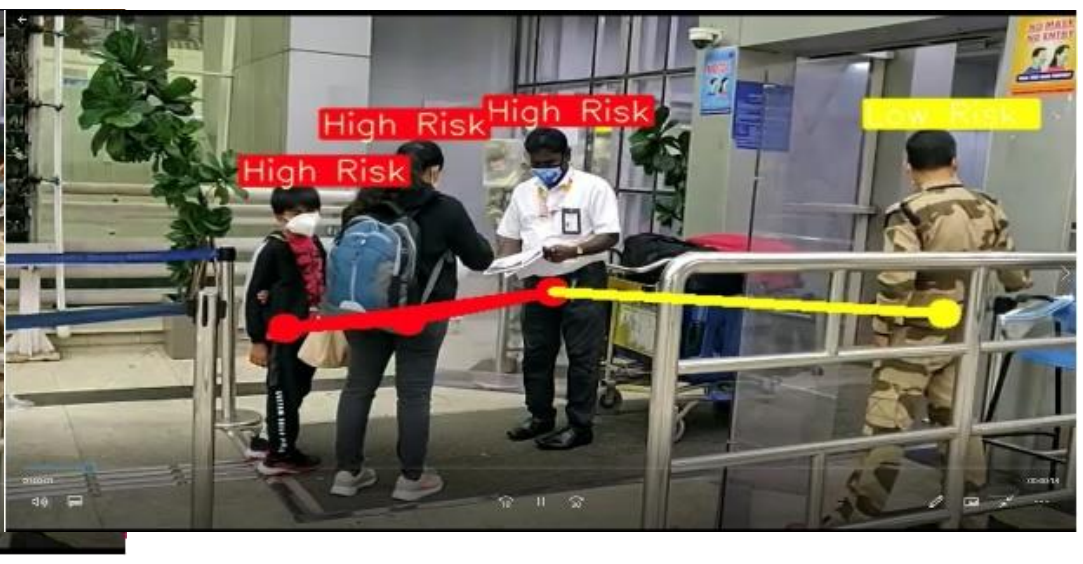

## 5.2 Part2: Comparison of YOLOv5s6 (default model trained for 300 epochs using pre-defined weights) vs YOLOv5s6 modified backbone CSP (<u>our custom model</u> trained for 300 epochs without using pre-defined weights) -- COCO 128 Dataset

The accuracy of 81.7% for the modified bottleneck CSP without transfer learning is observed on COCO dataset after training the model for 305 epochs whereas for the same epochs, the default YOLOv5 model is attaining 80.1% accuracy with transfer learning. This shows an improvement in accuracy by our modified bottleneck CSP model.

| Model | class | images | labels | size (pixels) | Precision |
|---|---|---|---|---|---|
| YOLOv5s6 | all | 5000 | 36335 | 640 | 80.1 |
| YOLOv5s6_modified_bottleneckCSP | all | 5000 | 36335 | 640 | 81.7 |

**Table 1.** This clearly proves that the custom model is performing better than the default model on the COCO dataset.

### 5.3 Part3: Comparison of YOLOv5s (default model trained for 30 epochs using pre-defined weights) vs YOLOv5s6 modified backbone CSP (our custom model trained for 30 epochs without using pre-defined weights) -- Visdrone Dataset

For the Visdrone dataset, we are able to achieve an accuracy of upto 56.5% for certain classes and especially an accuracy of ~40% for people and pedestrians with transfer learning using the default YOLOv5s model for 30 epochs. While the modified bottleneck CSP is able to perform slightly better than the default model with an accuracy score of upto 58.5% for certain classes and an accuracy of ~42.1% for people and pedestrians.

This clearly proves that the custom model is performing better than the default model on the Visdrone dataset.

All of the benchmarking tests and comparisons were conducted on the same hardware and software.

| Model | Size (pixels) | Class | Precision | Recall | mAPval 0.5:0.95 | mAPval 0.5 |
|---|---|---|---|---|---|---|
| **YOLOv5s** | 640 | All | 38.1 | 30.4 | 14.4 | 27.8 |
| | 640 | Pedestrian | 41.8 | 37.7 | 14.5 | 36.1 |
| | 640 | People | 40.4 | 31.6 | 8.6 | 27.9 |
| | 640 | Bicycle | 17.1 | 11.6 | 2.21 | 6.5 |
| | 640 | Car | 56.5 | 71.9 | 45.5 | 70.6 |
| | 640 | Van | 30.4 | 36.0 | 18.3 | 28.0 |
| | 640 | Truck | 34.3 | 29.2 | 13.8 | 23.9 |
| | 640 | Tricycle | 38.4 | 6.6 | 5.83 | 11.6 |
| | 640 | Awning-tricycle | 31.1 | 5.08 | 4.31 | 7.18 |
| | 640 | Bus | 45.9 | 34.5 | 19.3 | 32.1 |
| | 640 | Motor | 44.7 | 39.8 | 11.9 | 34.1 |
| **YOLOv5s6_modified_bottleneckCSP** | 640 | All | 39.3 | 30.8 | 14.9 | 27.7 |
| | 640 | Pedestrian | 42.8 | 37.4 | 14.0 | 36.0 |
| | 640 | People | 41.4 | 32.6 | 8.9 | 28.9 |
| | 640 | Bicycle | 18.1 | 12.6 | 3.51 | 6.5 |
| | 640 | Car | 58.5 | 78.9 | 49.5 | 72.6 |
| | 640 | Van | 33.9 | 35.4 | 18.8 | 28.5 |
| | 640 | Truck | 34.7 | 30.4 | 14.7 | 24.3 |
| | 640 | Tricycle | 39.4 | 6.8 | 5.83 | 10.6 |
| | 640 | Awning-tricycle | 32.1 | 5.18 | 4.21 | 8.18 |
| | 640 | Bus | 46.0 | 35.5 | 19.9 | 32.9 |
| | 640 | Motor | 44.9 | 40.0 | 12.5 | 34.6 |

**Table 2.** YOLOv5s and YOLOv5s6_modified_bottleneckCSP benchmark comparison

## 5.4 Visualization(s)

Sample Outputs using our Custom Model Trained and Evaluated on Visdrone Dataset:

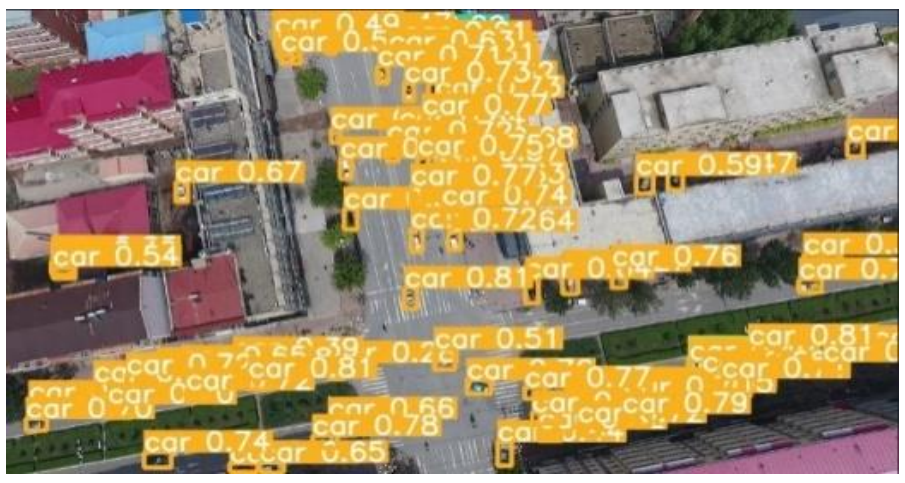

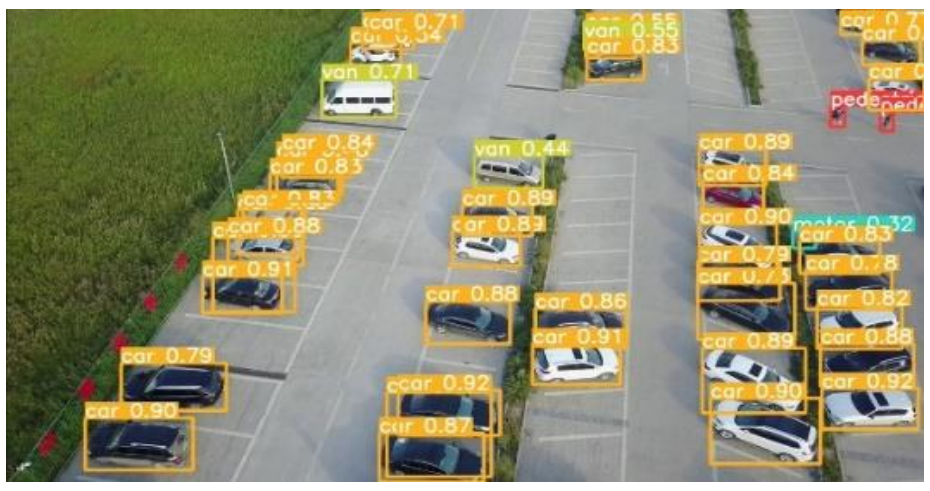

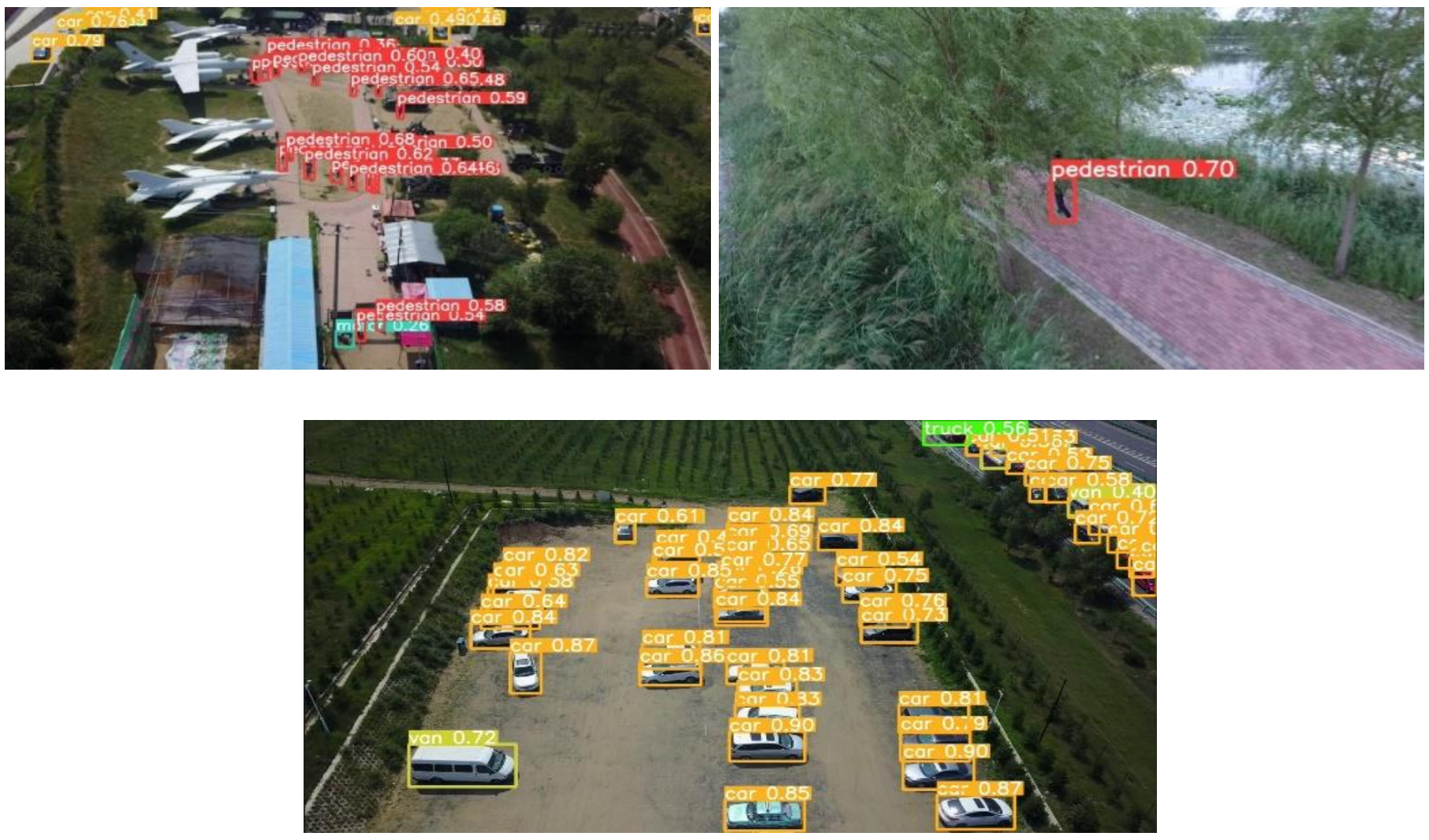

**Fig. 11.** Model Sample Outputs for Visdrone Dataset

## 6 FUTURE SCOPE AND CHALLENGES

We have learned to successfully train a model with and without predefined weights as well on different datasets thereby able to evaluate the model performance within a model as well as across models for both object detection and social distancing. Taking into account the importance of social distance in managing and reducing the probability of COVID-19 disease from continuously spreading which can cause the healthcare system to collapse due to high numbers of patients, our project can offer a smart solution to the public to monitor and remind them to maintain the distance when in public areas. Due to infrastructure limitation, we weren't able to train beyond a certain number of epochs which would have eventually helped us in getting better accuracy scores. In terms of data collection, we could partner with a few establishments which witness large footfalls such as supermarkets, malls, theatres, hospitals, banks, etc. and carry out the social distancing on a live feed and raise an alarm as when it is needed. In the future, additional backend processes will be included that allow advanced statistical analysis to be done which can be used by the authority, facilities or building owner to monitor the level of compliance among the people or visitors.

## 7 CONCLUSION

This research presented an intelligent surveillance system for social distancing classification. The proposed technique achieved promising results for people detection in terms of evaluating the accuracy and precision of the detector comparable to other YOLOv5 models. We have tested and concluded that the model performs well on side angle, top angle, overhead and drone angle. Any new angle other than the ones mentioned, the model may not work. The application is not intended to capture social distancing from Drone angle as the objects looks smaller and there would be significant overlaps in the prediction labels which might make the readability difficult to visualize.

As of now, the model is trained on separate datasets one by one but as an effective mechanism, we can consolidate different datasets thereby being able to train the model on multiple image variances which will help us achieve a better prediction on any dataset without needing to re-train. There may also be genuinely raised concerns about privacy and individual rights which can be addressed with some additional measures such as prior consents for such working environments, hiding a person's identity in general, and maintaining transparency about its fair uses within limited stakeholders. The proposed approach can be implemented in a distributed video surveillance system, Drone surveillance system and other similar surveillance systems. It is a suitable solution for the authorities to visualize the compliance of people with social distancing at a confidence of more than 80% with limited training itself. With a significant amount of training, the accuracy scores can reach above 90%.